\documentclass{article}

\RequirePackage{tikz}

\usepackage{arxiv}

\usepackage[utf8]{inputenc} 
\usepackage[T1]{fontenc}    
\usepackage{hyperref}       
\usepackage{url}            
\usepackage{booktabs}       
\usepackage{amsfonts}       
\usepackage{nicefrac}       
\usepackage{microtype}      
\usepackage[square,numbers]{natbib}
\usepackage{doi}

\usepackage{algorithm}
\usepackage{algpseudocode}

\usepackage{pgfplots}
\usepackage{pgfplotstable}
\pgfplotsset{compat=1.7}
\usepackage{tikz}
\usepackage{graphicx}
\usepackage{subcaption}

\title{NeuroCERIL: Robotic Imitation Learning via Hierarchical Cause-Effect Reasoning in Programmable Attractor Neural Networks}

\author{Gregory P. Davis \\
	Department of Computer Science\\
	University of Maryland\\
	College Park, MD, USA \\
	\texttt{grpdavis@umd.edu} \\
	\And
	Garrett E. Katz \\
	Department of Elec. Engr. and Comp. Sci.\\
	Syracuse University\\
	Syracuse, NY, USA \\
	\texttt{gkatz01@syr.edu} \\
	\And
	Rodolphe J. Gentili \\
	Department of Kinesiology\\
	University of Maryland\\
	College Park, MD, USA \\
	\texttt{rodolphe@umd.edu} \\
	\And
	James A. Reggia \\
	Department of Computer Science\\
	University of Maryland\\
	College Park, MD, USA \\
	\texttt{reggia@umd.edu} \\
}

\renewcommand{\headeright}{Preprint}
\renewcommand{\undertitle}{Preprint}
\renewcommand{\shorttitle}{NeuroCERIL}

\hypersetup{
pdftitle={NeuroCERIL: Robotic Imitation Learning via Hierarchical Cause-Effect Reasoning in Programmable Attractor Neural Networks},
pdfsubject={cs.AI},
pdfauthor={Gregory P. ~Davis, Garrett E. ~Katz, Rodolphe J. ~Gentili, James A. ~Reggia},
pdfkeywords={Imitation learning, Causal reasoning, Programmable neural networks, Cognitive control, Working memory, Symbolic processing},
}

\begin{document}
\maketitle

{
  \centerline
  {\large \bfseries \scshape Abstract}
  \begin{quote}
Imitation learning allows social robots to learn new skills from human teachers without substantial manual programming, but it is difficult for robotic imitation learning systems to generalize demonstrated skills as well as human learners do. Contemporary neurocomputational approaches to imitation learning achieve limited generalization at the cost of data-intensive training, and often produce opaque models that are difficult to understand and debug. In this study, we explore the viability of developing purely-neural controllers for social robots that learn to imitate by reasoning about the underlying intentions of demonstrated behaviors. We present NeuroCERIL, a brain-inspired neurocognitive architecture that uses a novel hypothetico-deductive reasoning procedure to produce generalizable and human-readable explanations for demonstrated behavior. This approach combines bottom-up abductive inference with top-down predictive verification, and captures important aspects of human causal reasoning that are relevant to a broad range of cognitive domains. Our empirical results demonstrate that NeuroCERIL can learn various procedural skills in a simulated robotic imitation learning domain. We also show that its causal reasoning procedure is computationally efficient, and that its memory use is dominated by highly transient short-term memories, much like human working memory. We conclude that NeuroCERIL is a viable neural model of human-like imitation learning that can improve human-robot collaboration and contribute to investigations of the neurocomputational basis of human cognition.
  \end{quote}
}

\keywords{Imitation learning \and Causal reasoning \and Programmable neural networks \and Cognitive control \and Working memory \and Symbolic processing}

\maketitle

\section{Introduction}\label{sec1}

Humans readily teach and learn using demonstration and imitation.
The ability to imitate emerges at an early age and plays a crucial role in early human development, but remains a natural and intuitive method for acquiring new skills throughout the lifespan \citep{jones2009development, meltzoff2009foundations}.
Crucially, human-level imitation involves not only replicating observable motor behavior, but also inferring the underlying goals and intentions of the demonstrator.
This allows learners to generalize demonstrated skills to novel environments by abstracting away details that are circumstantial to the demonstration environment.

Programming social robots to carry out complex tasks in a human-like fashion is difficult and typically requires laborious programming by an experienced roboticist.
One promising solution to this problem is to develop robots that learn from demonstrations (i.e., robotic imitation learning) \citep{ravichandar2020recent, hussein2017imitation, billard2008survey, schaal1999imitation}.
This eases the burden of robotic programming, making it accessible to non-experts.
However, most work in robotic imitation learning focuses on reproducing overt motor activity, which affords only limited generalization \citep{ravichandar2020recent}.
Developing more human-like imitation in robots requires algorithms for reasoning about observed actions to construct a deeper understanding of the demonstrator's goals and intentions that can be adapted to novel environments.
This approach also provides a common framework for reasoning about human and robot behavior, which facilitates an understanding of roles and perspectives that promotes seamless human-robot collaboration \citep{trafton2005enabling}.


Inferring intentions during imitation learning can be viewed as a process of causal reasoning, in which observable behaviors are treated as the effects of hidden causal intentions.
Reasoning backwards from effects to possible causes is known as abductive inference, and is a crucial aspect of human diagnostic reasoning and general problem-solving.
Algorithms for causal imitation learning may therefore be broadly relevant to cognitive domains beyond overt motor planning, such as language comprehension and visual scene understanding.
In addition, insight into the neural basis of these cognitive processes may be gained through the development of robotic imitation learning systems that respect the constraints and neurobiological foundations of human cognition.




In previous work, we developed CERIL, a robotic imitation learning system that uses abductive inference to construct causal interpretations of demonstrated motor behavior and generalizes them for imitation in novel environments \citep{katz2017novel}.
While effective and provably correct, CERIL's algorithms are implemented with traditional non-neural symbolic programming and have a limited degree of cognitive plausibility.
Specifically, CERIL's inference algorithm involves exhaustive enumeration of plausible causal explanations, which places unrealistic demands on working memory.
It also processes demonstrations in an offline fashion, rather than iteratively as humans do.

In this paper, we present NeuroCERIL, a purely-neural imitation learning system that reproduces CERIL's ability to explain demonstrated behavior during imitation learning. NeuroCERIL is a programmable neural network that implements a novel causal inference algorithm based on the hypothetico-deductive approach, combining bottom-up abductive inference with top-down deductive prediction and verification. Notably, NeuroCERIL processes demonstrations in an online fashion by iteratively constructing efficient data structures in memory that can be used to generate plausible explanations for observed behavior.
In other words, NeuroCERIL's cognitive processes are much more human-like than CERIL's, and they are supported by neurocomputational mechanisms that more closely resemble those used by people during cause-effect reasoning.

NeuroCERIL's neural architecture is an extension of NeuroLISP, a neural interpreter for a subset of the Common LISP programming language \citep{davis2022neurolisp}, but includes two major additions: a class system for constructing typed objects with named attributes, and an exception handling system for responding to errors encountered during program evaluation.
These innovations show how high-level programming constructs can improve the efficiency and computational capabilities of programmable neural networks, allowing them to learn complex cognitive behaviors.

Our empirical results show that NeuroCERIL is potentially an effective neurocognitive controller for robotic imitation learning systems, as it is able to reproduce CERIL's performance on a battery of demonstrations of procedural maintenance tasks. We examine NeuroCERIL's runtime and memory usage during causal inference and show that they scale roughly linearly with the length of the demonstration. Further, many of its memories have very short lifetimes, and are only accessed during a narrow window of processing. Thus, like human working memory, many of its short-term memories are rapidly abandoned, and only a small fraction of its memories need to be maintained through the duration of a demonstration.

\section{Related Work}\label{sec2}

In imitation learning, or learning from demonstration, an agent learns new skills by observing a teacher's demonstrations.
While imitation can be as simple as reproducing motor trajectories, humans are capable of a higher form of imitation that involves reasoning about a teacher's goals and intentions.
This form of imitation, which we refer to as ``cognitive-level" imitation, allows learners to grasp the underlying purpose of the demonstrated behaviors and generalize them to novel circumstances.
Although its origin remains unclear, cognitive-level imitation emerges in early childhood and plays a crucial role in human cognitive development \citep{bandura2017psychological, jones2009development, meltzoff1995understanding, baldwin2001discerning, tomasello1993cultural}.
Imitation is thought to be supported by neural mechanisms that establish shared representations for perceptually observable behavior and cognitive-motor control processes (i.e., the mirror neuron system), facilitating perspective-taking and interpersonal collaboration \citep{oztop2013mirror, jackson2006neural, fogassi2005parietal, koster2020motor}.

%
Robotic imitation learning has been proposed as a solution to the complexity and limited accessibility of robotic programming \citep{ravichandar2020recent, hussein2017imitation, billard2008survey, schaal1999imitation}.
Despite recent progress, it remains difficult to develop systems that generalize well to novel circumstances and adapt learned behavior to situations that deviate from the demonstration environment.
Furthermore, contemporary robotic imitation learning systems often rely on machine learning techniques that require substantial training data and are opaque to users, which makes it difficult to diagnose and debug errors, and creates barriers in trustworthiness and explainability.
Safe and effective robotic imitation learning requires human-like algorithms for understanding demonstrated actions, adapting learned skills to novel environments, and constructing explanations of planned behavior that can be understood by end-users.

\begin{figure*}[!b]
\begin{center}
\includegraphics[width=0.7\textwidth]{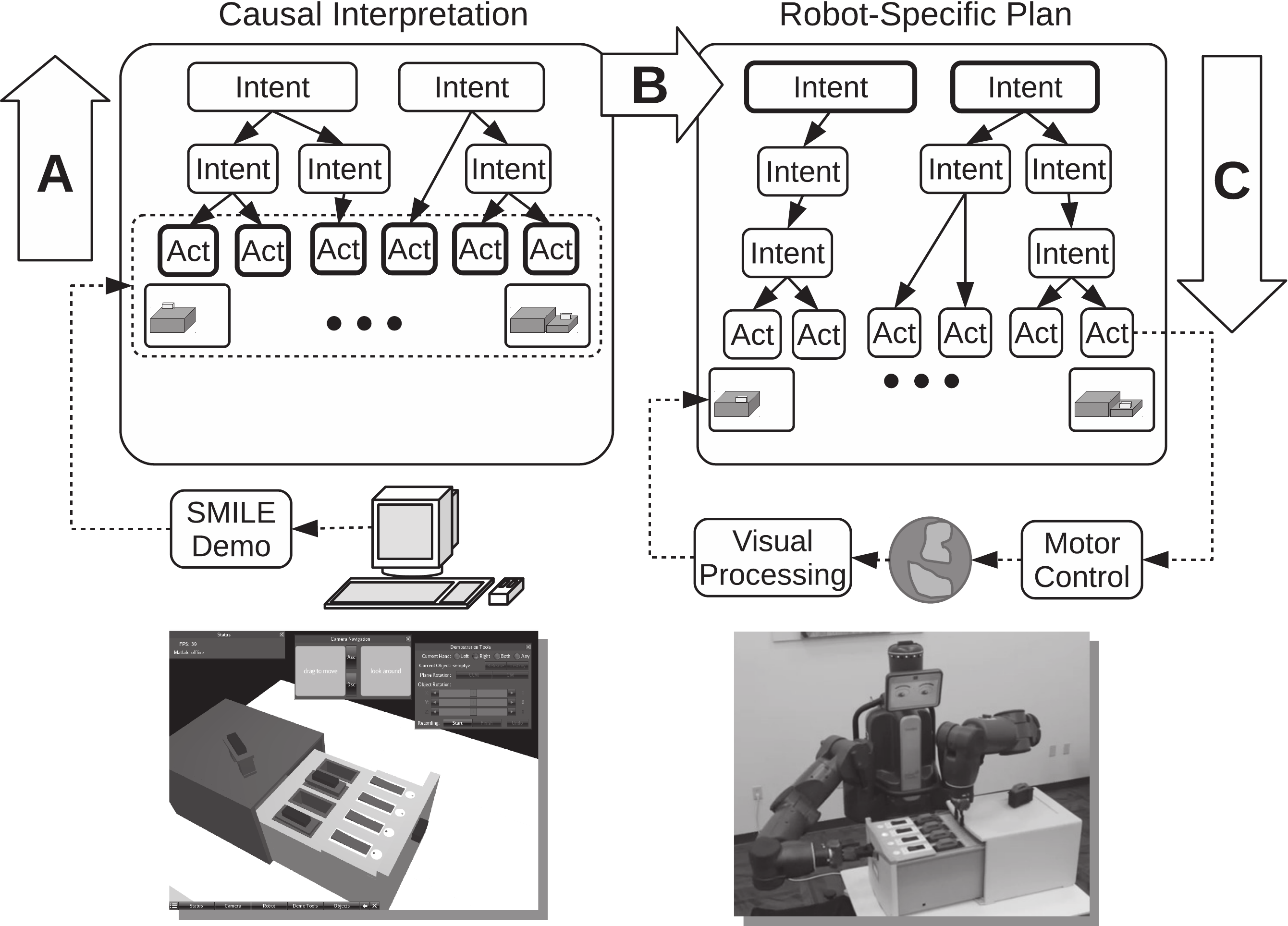}
\end{center}
\caption{ Overview of CERIL's cause-effect reasoning framework for imitation learning. Demonstrations of a procedural skill are recorded in a virtual environment (SMILE, bottom left) and interpreted to construct a hierarchical causal explanation for observed behavior (left side, \textbf{A}). This explanation is then adapted during imitation to construct a robot-specific plan that implements the learned skill (right side, \textbf{B} and \textbf{C}). CERIL issues low-level motor commands that manipulate objects identified in the environment using visual processing (bottom right). In the present work, we introduce NeuroCERIL, a neurocomputational model that performs the causal interpretation process on the left side of the figure using human-like hypothetico-deductive reasoning. Figure reproduced from \cite{katz2017novel}. }\label{fig:ceril}
\end{figure*}

In previous work, we addressed these challenges by developing CERIL, a robotic imitation learning system based on cause-effect reasoning (Figure \ref{fig:ceril}) \citep{katz2017novel, reggia2018humanoid}.
The basic intuition behind this approach is that demonstrated motor behavior is caused by hidden intentions or goals that must be inferred by the learner.
To do so, CERIL reasons backwards from effects to plausible causes, a process known as abductive inference, to construct a hierarchy of cause-effect relations that explains the demonstration in terms of high-level intentions.
Because these intentions are abstracted from the concrete demonstration, they can be used during imitation to plan a new sequence of motor behavior that implements the learned skill in a novel environment.
In addition, CERIL can use these plans to provide explanations for its motor behavior, which allows end-users to investigate and debug its understanding of demonstrated skills.
Thus, CERIL uses cause-effect reasoning to understand demonstrated behavior, generalize it to new environments, and explain its own behavior to human users.

CERIL learns from demonstrations that take place in SMILE, a virtual 3D environment with a tabletop and various objects that can be manipulated through manual interactions, such as grasping and moving a block or toggling a switch \citep{huang2015virtual}.
Users simulate a sequence of actions demonstrating a skill to be learned, and SMILE records the sequence along with changes in the state of the environment, such as object locations and mutable properties (e.g., the state of a toggle switch).
A transcript of the recorded demonstration is then passed to CERIL, which generates an explanation composed of hidden causal intentions that are inferred using a preprogrammed knowledge-base of cause-effect relations.



As noted above, CERIL is effective and provably correct, but its algorithms are implemented with traditional non-neural programming techniques that have a limited degree of cognitive plausibility.
To infer intentions, CERIL uses a bottom-up dynamic programming algorithm that exhaustively enumerates plausible causal explanations, which places unrealistic demands on working memory.
It also requires multiple passes through a demonstration, whereas human imitators reason about demonstrated behavior as it occurs to construct partial explanations before a demonstration is complete.
Finally, it is unclear how this approach might be implemented using neural networks to leverage the unique advantages of neural computation, such as its capacity for learning and generalization, and provide insight into the neurobiological foundations of human imitation learning.



Neural approaches to imitation learning typically involve deep neural networks, which require data-intensive training and exhibit a limited degree of generalization in constrained environments \citep{duan2017one, liu2018imitation, xu2018neural, sun2018neural}.
It remains an outstanding challenge to develop neural networks with the high-level cognitive abilities that CERIL exhibits, such as causal inference, compositional modeling, and logical reasoning.
Many neural models are incorporated into hybrid systems that delegate these abilities to non-neural algorithms, such as neural-guided search and program synthesis \citep{bunel2018leveraging, silver2016mastering, kalyan2018neural}.
Furthermore, these models often rely on neurobiologically implausible processes that require simultaneous access to a temporal sequence of inputs or activity states, such as temporal convolutions and attention.

In recent work, we have developed neurocognitive systems that learn to represent and evaluate symbolic programs (i.e., programmable neural networks) \citep{davis2021compositional, katz2019programmable, sylvester2016engineering}.
Most recently, we developed NeuroLISP, a programmable neural network with a compositional working memory that can learn to evaluate programs written in LISP, a language with an extensive history in artificial intelligence research \citep{davis2022neurolisp}.
Several features make NeuroLISP an attractive option for modeling human-level cognition in a neurobiologically plausible manner.
NeuroLISP can learn to perform high-level cognitive tasks that are difficult for contemporary deep neural networks, such as compositional sequence manipulation, tree traversals, and symbolic pattern matching.
Importantly, it learns these tasks using fast associative learning rules that establish robust algorithmic behavior using minimal training data.
Its architecture is program-independent, and it represents programs and data using learned attractor states in recurrent neural regions (i.e., distributed representations) that are controlled by top-down gating of both learning and activation.
Finally, programmable neural networks can be integrated with neural models of sensorimotor control, including the complex motor control required in robotic imitation learning \citep{katz2021tunable, gentili2015neural}.



In this paper, we explore whether it is viable to develop purely-neural controllers for social robotic systems that behave in a human-like manner.
To this end, we develop NeuroCERIL, a programmable neural network that learns human-like algorithms for causal inference during imitation learning (left side of Figure \ref{fig:ceril}).
We evaluate our model using CERIL as a target system, as it has been demonstrated to be an effective cognitive controller for bimanual robots.
To address CERIL's limitations in cognitive plausibility, NeuroCERIL implements a novel causal inference algorithm based on the hypothetico-deductive approach, an influential model of diagnostic and scientific reasoning \citep{lawson2000humans, sprenger2011hypothetico, marcum2012integrated, reggia1987modeling, lawson2000generality}.
Hypothetico-deductive reasoning involves a combination of bottom-up abductive inference and top-down predictive verification, which obviates the need for exhaustive search by focusing cognitive processing on relevant causal knowledge.
NeuroCERIL therefore serves as a purely-neural controller for social robots that more closely resembles human cognition and learning, facilitates more seamless human-robot interactions, and provides a framework for modeling cognitive processes that are relevant to a broad range of application domains.

\section{Methods}\label{sec3}

NeuroCERIL\footnote{https://github.com/vicariousgreg/neuroceril} is a brain-inspired cognitive model that learns procedural skills from demonstrations using cause-effect reasoning.
The model's architecture is an extension of NeuroLISP, a programmable neural network that can store and evaluate programs written in a subset of the Common LISP programming language \citep{davis2022neurolisp}.
NeuroCERIL is programmed with a novel causal inference algorithm based on hypothetico-deductive reasoning, which combines bottom-up abductive inference with top-down deductive prediction and verification.
This approach allows NeuroCERIL to anticipate future behavior and focuses cognitive computations on plausible explanations for observed behavior.

Although NeuroCERIL is implemented using attractor neural networks, its distributed neural computations represent algorithmic procedures performed on symbolic data structures.
It is therefore convenient to begin by describing its behavior in terms of symbolic information processing.
We first outline the robotic imitation learning domain in which NeuroCERIL operates (Section \ref{sec3.1}), and the algorithms and data structures that it uses to implement hypothetico-deductive causal inference (Section \ref{sec3.2}).
Then, we present the neurocognitive architecture that learns to represent and evaluate these algorithms and data structures using only neural computations (Section \ref{sec3.3}). Finally, we describe the empirical experiments that we conducted to validate NeuroCERIL, including a battery of test demonstrations that was used to validate CERIL (Section \ref{sec3.4}).
We show that NeuroCERIL performs comparably to CERIL, but that its iterative hypothetico-deductive approach is memory efficient and scales well to long demonstrations.



\subsection{Robotic Imitation Learning Domain}\label{sec3.1}

\begin{figure*}[!b]
\begin{center}
\includegraphics[width=0.85\textwidth]{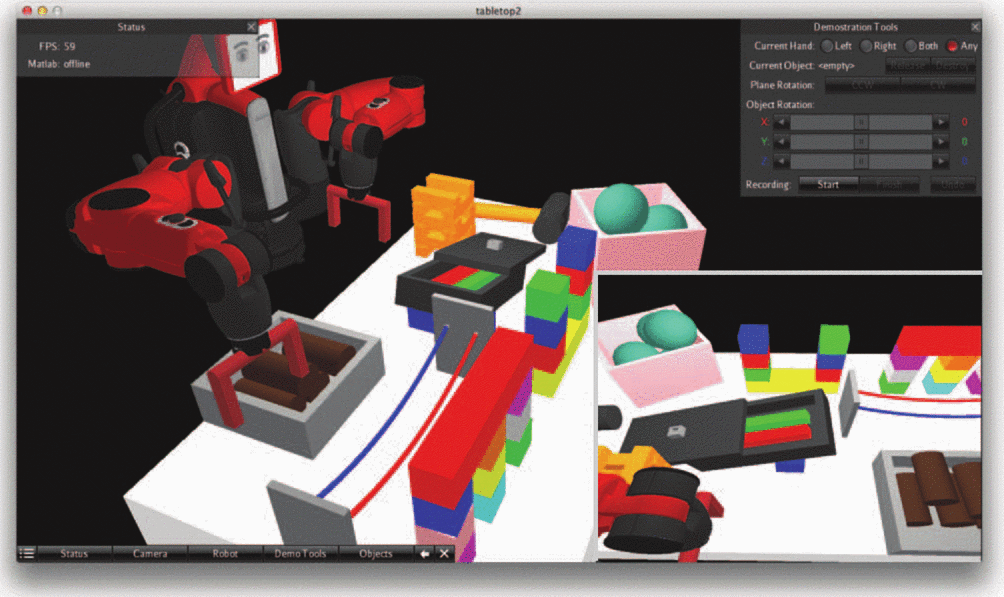}
\end{center}
\caption{ Virtual 3D environment used for recording procedural demonstrations for imitation learning (SMILE: Simulator for Maryland Imitation Learning Environment). The environment contains a tabletop with various objects that can be manipulated, as well as an avatar for a bimanual robot, Baxter (Rethink Robotics). The user can pick up, rotate, move, and release objects on the tabletop to create a sequential demonstration to be used for imitation learning. Figure reproduced from \cite{huang2015virtual}. }\label{fig:smile}
\end{figure*}

NeuroCERIL operates in the robotic imitation learning domain designed for CERIL, which involves a bimanual robot (Baxter) learning procedural maintenance tasks \citep{katz2017novel}.
As previously mentioned, a teacher demonstrates these tasks using SMILE, a simulated 3D environment that allows users to interact with virtual objects such as blocks, drawers, switches, and screw valves \citep{huang2015virtual}.
SMILE also includes a simulation of the Baxter robot, shown in Figure \ref{fig:smile} with a variety of simulated objects.
SMILE greatly simplifies the low-level sensory processing involved in recognizing and segmenting actions and objects, allowing us to focus on the higher level cognitive processing that occurs during imitation.
When a user is finished recording a demonstration, SMILE produces a transcript containing the sequence of recorded actions, along with a record of changes that occur in the environment, such as changes in object state or location.

Actions are encoded as discrete events with free parameters that refer to objects or locations in the environment.
For example, grasping a \texttt{red-block} with the \texttt{left-gripper} is encoded as:
\begin{quote}
    \texttt{grasp<red-block, left-gripper>}
\end{quote}
\noindent The identifier \texttt{left-gripper} refers to the demonstrator's left hand, and \texttt{red-block} refers to an object in the environment, which is encoded as a collection of named properties:
\begin{quote}
    \texttt{\{id:red-block, type:block, color:red, location:loc\}}
\end{quote}
\noindent Once the block is grasped, its \texttt{location} property is updated to \texttt{left-gripper} to indicate that it is currently located in the demonstrator's left hand. This change is represented as a record containing the object identifier, property name, and the new property value:
\begin{quote}
    \texttt{(red-block location left-gripper)}
\end{quote}
Once the block is moved and placed, this property is updated again to reflect its new location. Although locations are encoded as discrete symbols, they can be associated with representations of 3D points in continuous space for use in low-level motor planning.


Like CERIL, NeuroCERIL is pre-programmed with a knowledge-base of cause-effect relations that describe the implementation of abstract intentions.
These causal relations are used during learning to infer a demonstrator's intentions (goals, on left side of arrow) from the demonstrator's actions (right side of arrow).
For example, the intention to \texttt{relocate} an object (\texttt{obj}) to a target location (\texttt{loc}) causes a sequence of concrete motor actions: \texttt{grasp} the object, \texttt{move} it to a target location, and \texttt{release} the grasp.
This is encoded as a template or schema that can be matched to observed behavior:
\begin{quote}
    \texttt{relocate<obj,loc>} $\rightarrow$
    \begin{quote}
    \texttt{grasp<obj, gripper>, \\ move<gripper, loc>, \\ release<gripper>}
    \end{quote}
\end{quote}
\noindent Here, the right arrow represents causation, and indicates that the intention on the left side of the arrow can cause the ordered sequence of actions on the right side.
It is important that parameter names (\texttt{obj}, \texttt{loc}, \texttt{gripper}) are repeated in this schema, because this indicates correspondences between the parameters of the intention and the actions that it causes (e.g., the same gripper is used for each action).
NeuroCERIL verifies these correspondences when it infers casual intentions in a demonstration.
In addition, each schema may include explicit logical predicates that must be satisfied for a cause-effect relation to be plausible.
For example, the intention to open a drawer may cause a sequence of grasping, moving, and releasing, but the grasped object must be a drawer handle, and the drawer must be closed prior to opening.
These constraints can be encoded as logical statements that NeuroCERIL must verify while inferring causal intentions.

The effects of a causal intention may include other abstract intentions, allowing causes to be chained together to create hierarchies of cause-effect relations.
For example, the intention to swap the location of two objects may be implemented as a sequence of \texttt{relocate} intentions:
\begin{quote}
    \texttt{swap<obj1, obj2>} $\rightarrow$ 
    \begin{quote}
        \texttt{relocate<obj1, temp>, \\ relocate<obj2, loc1>, \\ relocate<obj1, loc2>}
    \end{quote}
\end{quote}
\noindent A concrete demonstration of this swapping behavior would involve a sequence of \texttt{grasp}, \texttt{move}, and \texttt{release} actions that are caused by intermediate \texttt{relocate} intentions.
Thus, inferring the causes of demonstrated actions requires a recursive inference process: when an intention is recognized as a plausible cause, it is treated as the effect of plausible higher-level causal intentions.

NeuroCERIL's causal knowledge-base may contain multiple schemas describing different implementations of the same intention.
For example, the location of two objects may be swapped without placing one in an intermediate location, by instead keeping one object in hand while relocating the other:
\begin{quote}
    \texttt{swap<obj1, obj2>} $\rightarrow$
    \begin{quote}
        \texttt{grasp<obj1, gripper>, \\ move<gripper, temp>, \\ relocate<obj2, loc1>, \\ move<gripper, loc2>, \\ release<gripper>}
    \end{quote}
\end{quote}
\noindent Here, \texttt{temp} refers to a location in the air to which the demonstrator lifts \texttt{obj1} to, holding it there while \texttt{obj2} is relocated to the original location of \texttt{obj1} (\texttt{loc1}).
A key feature of NeuroCERIL's knowledge-base is that causal relations are agnostic to the implementation of their effects: a higher-level intention that is implemented using \texttt{swap} does not specify which implementation of \texttt{swap} to use.
This flexibility affords generalization during imitation; a demonstration involving one implementation of \texttt{swap} can be imitated using the other implementation.
Thus, causal inference allows the imitator to abstract away circumstantial details of the demonstration environment and adapt learned skills to novel circumstances.
This may also be necessary if the embodiment of the imitator differs from that of the demonstrator (e.g., number of arms, dexterity, range of motion), requiring the imitator to implement learned skills in a different but equivalent way.

Finally, a sequence of demonstrated actions may have more than one plausible explanation.
This may occur if two sequences of cause-effect relations share the same sequence of effects.
For example, given the following three cause-effect relations (shown without parameters for simplicity):
\begin{quote}
    \texttt{X $\rightarrow$ A, B} \\
    \texttt{Y $\rightarrow$ C} \\
    \texttt{Z $\rightarrow$ A, B, C}
\end{quote}

\noindent a sequence of actions \texttt{(A, B, C)} may be caused by the sequence of intentions \texttt{(X, Y)}, or the single intention \texttt{Z}.
In this case, the most parsimonious (i.e., simplest or shortest) explanation is usually preferred: \texttt{(A, B, C)} was caused by \texttt{Z}.

In the next subsection, we describe the new hypothetico-deductive causal inference algorithm that NeuroCERIL uses to identify the most parsimonious explanation for a demonstration.
NeuroCERIL is provided with a pre-programmed knowledge-base of cause-effect relations with optional logical constraints, as described above.
The initial state of the virtual environment is provided as a list of objects encoded as collections of named properties, which may change during the demonstration (e.g., location).
The demonstration is encoded as a sequence of parameterized actions, each paired with a list of changes that occur to objects in the environment.
The output of this algorithm is a sequence of top-level intentions identified as causes of the demonstrated actions, which serves as an explanation of the demonstration as well as an encoding of the demonstrated skill.

\subsection{Hypothetico-Deductive Causal Inference}\label{sec3.2}

\begin{figure}[!t]
\begin{center}
\includegraphics[width=0.65\columnwidth]{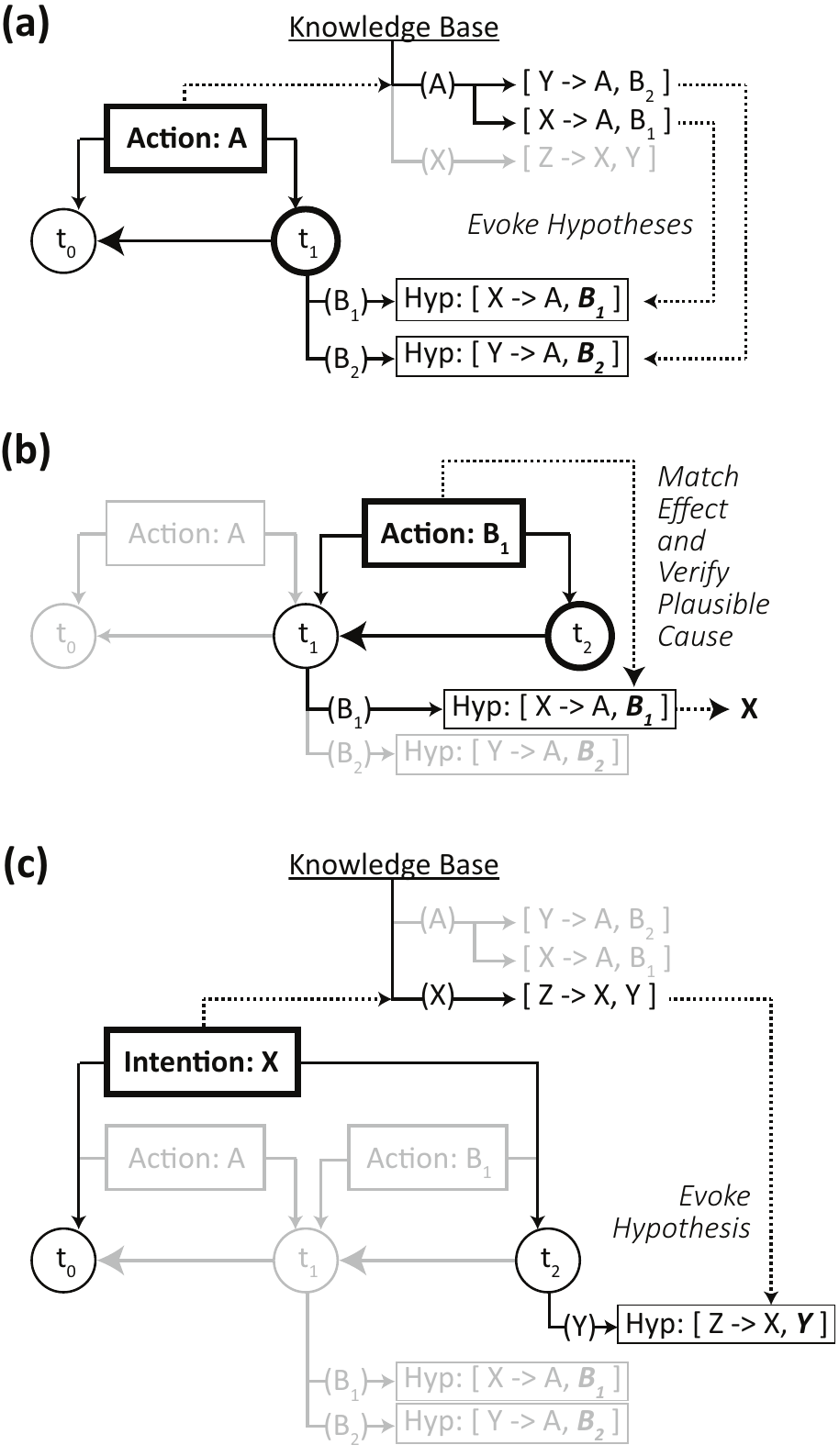}
\end{center}
\caption{ Hypothetico-deductive process for inferring hierarchical intentions during imitation learning. \textbf{(a)} An action of type A is observed at timepoint $t_1$ and added to a chain of timepoints in memory (bold box and circles, left). The knowledge base (top right) is consulted to evoke plausible hypotheses about the action's cause. These hypotheses are added to timepoint $t_1$ (bottom right) and stored according to their subsequent predictions (B$_1$ and B$_2$). \textbf{(b)} An action of type B$_1$ is observed at $t_2$ (center), and matched to the hypothesis that predicted it, generating a plausible causal intention of type X (bottom right). \textbf{(c)} This intention is processed recursively as an observation spanning $t_0$ to $t_2$ (left). A hypothesis is evoked that predicts an intention of type Y at $t_2$ (bottom right).
}\label{fig:timeline}
\end{figure}

NeuroCERIL's approach to causal inference differs from CERIL's in a way that is more cognitively plausible and memory efficient.
Whereas CERIL conducts an exhaustive bottom-up search that makes multiple passes through an entire demonstration, NeuroCERIL uses a more human-like hypothetico-deductive approach that involves a combination of bottom-up and top-down reasoning to iteratively construct a causal explanation for a demonstration as it occurs.
When an action is observed, NeuroCERIL consults its cause-effect knowledge-base to generate explicit hypotheses about the demonstrator's causal intentions (bottom-up), and uses them to deduce testable predictions about subsequent actions (top-down).
NeuroCERIL's cognitive processing is focused on evaluating these predictions to verify or falsify hypotheses.
By organizing active hypotheses based on their predictions, NeuroCERIL can efficiently access those that are relevant to an observation, and avoid considering those that are not.
When all of the predictions of a hypothesis are verified by observations, the hypothesized causal intention is treated as an observation and processed recursively to identify plausible higher-level intentions that may have caused it.
In this way, NeuroCERIL constructs hierarchies of cause-effect relations that are supported by observations, and that represent plausible explanations for sequences of demonstrated behavior.
As plausible intentions are identified, NeuroCERIL updates \textit{parsimony pointers} that indicate the shortest sequence of intentions that covers the actions observed so far.
At the end of the demonstration, these pointers are traced back to identify the most parsimonious explanation for the entire demonstration.
This process is illustrated in Figure \ref{fig:timeline}, outlined as pseudocode in Algorithm \ref{alg:pseudocode}, and described in more detail below.

\begin{algorithm*}
\begin{algorithmic}
\Procedure {explain}{$demo$, $init\_env$} \Comment {infers causal explanation for $demo$nstration}
   \State $curr\_env$ $\gets$ $init\_env$ \Comment {initial environment state}
   \State $prev\_time$ $\gets$ create timepoint with $init\_env$ \Comment {initial timepoint}
   \For {each $action$ and record of environment $changes$ in $demo$}
      \State $curr\_env$ $\gets$ new environment with $changes$ chained off prior $curr\_env$
      \State $curr\_time$ $\gets$  new timepoint with $curr\_env$ chained off $prev\_time$
      \State set $action$ start and end timepoints to $prev\_time$ and $curr\_time$
      \State \Call {process\_action}{$action$}
      \State $prev\_time$ $\gets$ $curr\_time$
   \EndFor
   \State \Return \Call {trace} {$curr\_time$} \Comment {reconstruct top-level explanation from timeline}
\EndProcedure
\State
\Procedure {process\_action} {$action$} \Comment {updates timeline and hypotheses}
   \If {$action$ has shorter path to initial timepoint} \Comment{compare with parsimony pointer}
      \State update parsimony pointer for $action$'s end timepoint
   \EndIf
   \For {each $schema$ predicting $action$ type as first effect} \Comment{new hypotheses}
      \State $hypothesis$ $\gets$ generate hypothesis from $schema$ \Comment{abductive evocation}
      \State \Call {verify\_hypothesis}{$hypothesis,action$}
   \EndFor
   \For {each $hypothesis$ predicting $action$ type at $action$ end timepoint} \Comment{old hypotheses}
      \State \Call {verify\_hypothesis}{$hypothesis,action$}
   \EndFor
\EndProcedure
\State
\Procedure {verify\_hypothesis} {$hypothesis, action$} \Comment {verifies a $hypothesis$ for cause of $action$}
   \If {$action$ matches $hypothesis$ prediction} \Comment{unification and constraint checking}
      \If {$hypothesis$ is fully matched} \Comment {all predictions verified}
         \State $intent$ $\gets$ generate causal intention from $hypothesis$
         \State \Call {process\_action}{$intent$} \Comment{process inferred intention as an observed action}
      \Else
         \State update $hypothesis$ prediction
         \State add $hypothesis$ to $action$'s end timepoint \Comment{advance hypothesis in timeline}
      \EndIf 
   \EndIf
\EndProcedure
\State %
\Procedure {trace}{$curr\_time$} \Comment traces top-level explanation using parsimony pointers
   \State $intent$ $\gets$ parsimony pointer of $curr\_time$
   \State $prev\_time$ $\gets$ start timepoint of $intent$
   \If {$prev\_time$ is initial timepoint}
      \State \Return list containing $intent$ \Comment {first intention in top-level sequence}
   \Else
      \State $prior\_intents$ $\gets$ \Call {trace}{$prev\_time$} \Comment {recursion}
      \State append $intents$ to $prior\_intents$
      \State \Return $prior\_intents$
   \EndIf
\EndProcedure
\end{algorithmic}
\caption{Pseudocode for hypothetico-deductive causal inference algorithm}
\label{alg:pseudocode}
\end{algorithm*}

NeuroCERIL uses several different data structures to keep track of observed actions, their relative timing, and hypotheses about their causal explanations.
These data structures are organized around a timeline, represented in memory as a chain of discrete time-points that delimit observed actions (circles connected by solid arrows in Figure \ref{fig:timeline}).
Each action contains pointers to the timepoints immediately before and after it (i.e., start and end points).
For concrete primitive actions that are directly observed (e.g., grasping and releasing), these timepoints are adjacent in the timeline (\textbf{Action: A} in Figure \ref{fig:timeline}a).
However, inferred high-level causal intentions can be implemented with multiple lower-level actions, and can therefore span several timepoints (\textbf{Intention: X} in Figure \ref{fig:timeline}c).



Hypotheses originate from a bottom-up abductive reasoning process that we call \textit{evocation}; when an action/intention is observed, NeuroCERIL consults its causal knowledge-base to identify relevant cause-effect schemas that might explain it (top right of Figures \ref{fig:timeline}a and \ref{fig:timeline}c, and first loop of \textsc{process\_action} procedure in Algorithm \ref{alg:pseudocode}).
These schemas are stored in an associative array that maps each action/intention type to a list of schemas that predict it as their first effect.
For example, the knowledge-base may contain a schema describing a cause-effect relation between the \texttt{relocate} intention and a sequence of \texttt{grasp}, \texttt{move}, and \texttt{release} actions.
This schema is stored in the \texttt{grasp} list of the knowledge-base, and can be retrieved to evoke a hypothesis that an observed \texttt{grasp} action was caused by the intention to \texttt{relocate} the grasped object.
This hypothesis must be evaluated to determine if the observed action satisfies the constraints of the schema, including correspondences between parameters with shared names as well as explicit logical predicates that must be true for the causal relation to be plausible (\textsc{verify\_hypothesis} procedure in Algorithm \ref{alg:pseudocode}).
Corresponding parameters are matched with a symbolic pattern matching procedure (unification) that we have previously implemented using neural computations \citep{davis2022neurolisp}.
If these constraints are not satisfied, the hypothesis is immediately abandoned.
Otherwise, it is added to the timeline and used to make predictions about subsequent actions, as described below.
In Figure \ref{fig:timeline}a, the knowledge-base contains two schemas indicating causal relations that might explain the observed action of type A (top right).
The evoked hypotheses predict a subsequent action of type B$_1$ and B$_2$, respectively (bottom right).

Each timepoint contains a set of hypotheses that make predictions about actions or causal intentions that might occur immediately after it.
Like cause-effect schemas in the knowledge-base, these hypotheses are stored in an associative array that maps the predicted action/intention type to the hypotheses that predict it at that timepoint.
When an action/intention is observed, the hypothesis set for its starting timepoint is consulted to retrieve the hypotheses that predicted it (second loop of \textsc{process\_action} procedure in Algorithm \ref{alg:pseudocode}).
For the \texttt{relocate} example above, the evoked hypothesis predicts that the demonstrator will \texttt{move} the grasping arm immediately after the \texttt{grasp} action occurred.
When \texttt{move} is observed, this hypothesis is retrieved and evaluated to determine if its prediction was satisfied.
This involves verifying the schema's logical constraints, as described above.
If these constraints are satisfied and the hypothesis predicts further actions, the next prediction is retrieved, and the hypothesis is advanced to the next timepoint.
When all of the predictions for a hypothesis are verified, it is used to generate a plausible causal intention that is added to the timeline.
This intention is then processed recursively in order to generate and verify further hypotheses about its underlying cause (call to \textsc{process\_action} procedure within \textsc{verify\_hypothesis} procedure of Algorithm \ref{alg:pseudocode}).
In Figure \ref{fig:timeline}b, the observed action of type B$_1$ matches the prediction of a hypothesis at $t_1$, and the corresponding causal intention of type X is generated (bottom right).
This intention is then processed as an observation in Figure \ref{fig:timeline}c, and a new hypothesis is evoked proposing that an intention of type Z is the underlying cause.
This new hypothesis predicts an intention of type Y at time $t_2$, and is added to the timeline accordingly (bottom right).

\begin{figure*}[!b]
\begin{center}
\includegraphics[width=0.9\textwidth]{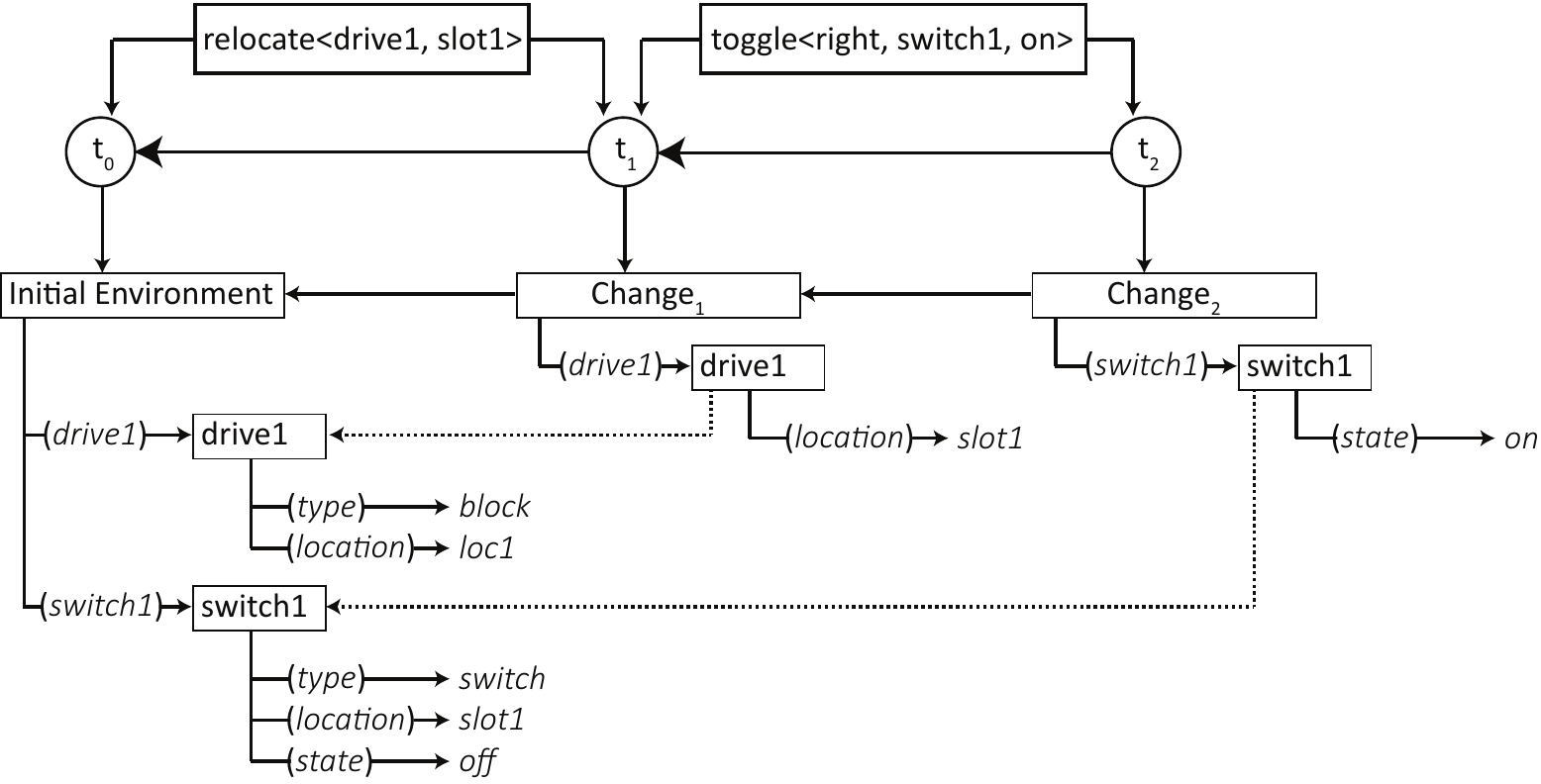}
\end{center}
\caption{ Representing a changing environment in memory during a simple demonstration, in which a hard drive (\textit{drive1}) is moved from an initial location (\textit{loc1}) to a slot (\textit{slot1}), and an adjacent switch (\textit{switch1}) is toggled \textit{on} with the \textit{right} hand. Each action is stored in a timeline of discrete timepoints (circles connected by solid arrows, top; see Figure \ref{fig:timeline}). The initial timepoint ($t_0$, top left) stores a representation of the initial environment as a nested associative array (``Initial Environment", left). Each entry in the array maps a symbolic name to an object (\textit{drive1} or \textit{switch1}, bottom left). Objects are stored as inner associative arrays, which map symbolic names of properties to their corresponding values (e.g., the \textit{location} of \textit{drive1} is initially \textit{loc1}). Subsequent timesteps store records of changes that occur in the environment (``Change$_1$" and ``Change$_2$", center and right). Like timepoints, these records are chained together in reverse chronological order, and each record is stored like the initial environment as a nested associative array. The inner associative arrays of corresponding objects are also chained together (dotted lines). This compact representation uses minimal memory, but affords access to the state of the environment at each timepoint.
}\label{fig:environment}
\end{figure*}

Each timepoint also contains a representation of the state of the environment that is consulted during hypothesis verification (Figure \ref{fig:environment}).
The environment contains several objects with named properties that can change over the course of a demonstration.
At the beginning of the demonstration, NeuroCERIL is provided with a full specification of the initial state of the environment.
When an action is observed, NeuroCERIL is also provided with a list of changes to object properties that were caused by the action.
Rather than maintaining full copies of the environment state at each timepoint, which would require substantial memory, NeuroCERIL stores a record of these changes that can be consulted to determine the state of the environment at a given timepoint (or its initial value if it was not changed).
To query the state of an object property at a given timepoint, NeuroCERIL retrieves the most recent change that occurred to that property prior to that timepoint.
To support this operation, each timepoint stores a nested associative array, where the outer array stores entries for changed objects, and each inner array stores entries for a particular object's changed properties.
Importantly, the inner arrays representing changes to the same object at different timepoints are chained together to allow an efficient search for the most recent change to a specific property (dotted lines in Figure \ref{fig:environment}).


Finally, NeuroCERIL maintains pointers in memory that can be used to retrieve the most parsimonious explanation for the actions observed so far.
The best explanation is the shortest sequence of intentions that covers all directly observed primitive actions without gaps or overlaps.
This is represented by a chain of alternating timepoints and intentions that leads from the last timepoint to the first timepoint.
Thus, each timepoint maintains a \textit{parsimony pointer} to the intention that provides the shortest path back to the first timepoint in the demonstration.
Whenever a plausible intention is identified, it is compared with the current best intention for the intention's end timepoint (beginning of \textsc{process\_action} procedure in Algorithm \ref{alg:pseudocode}).
NeuroCERIL performs this comparison by iterating through the paths simultaneously until the initial timepoint is reached.
If the newly identified intention provides a shorter path to the initial timepoint, it is replaced as the current best intention for the end timepoint.
When the demonstration is complete, the best explanation for the full sequence of observed actions can be reconstructed by following the chain of parsimony pointers from the final timepoint back to the initial timepoint (\textsc{trace} procedure in Algorithm \ref{alg:pseudocode}).

\subsection{Neural Implementation}\label{sec3.3}

\begin{figure*}[!t]
\begin{center}
\includegraphics[width=0.7\textwidth]{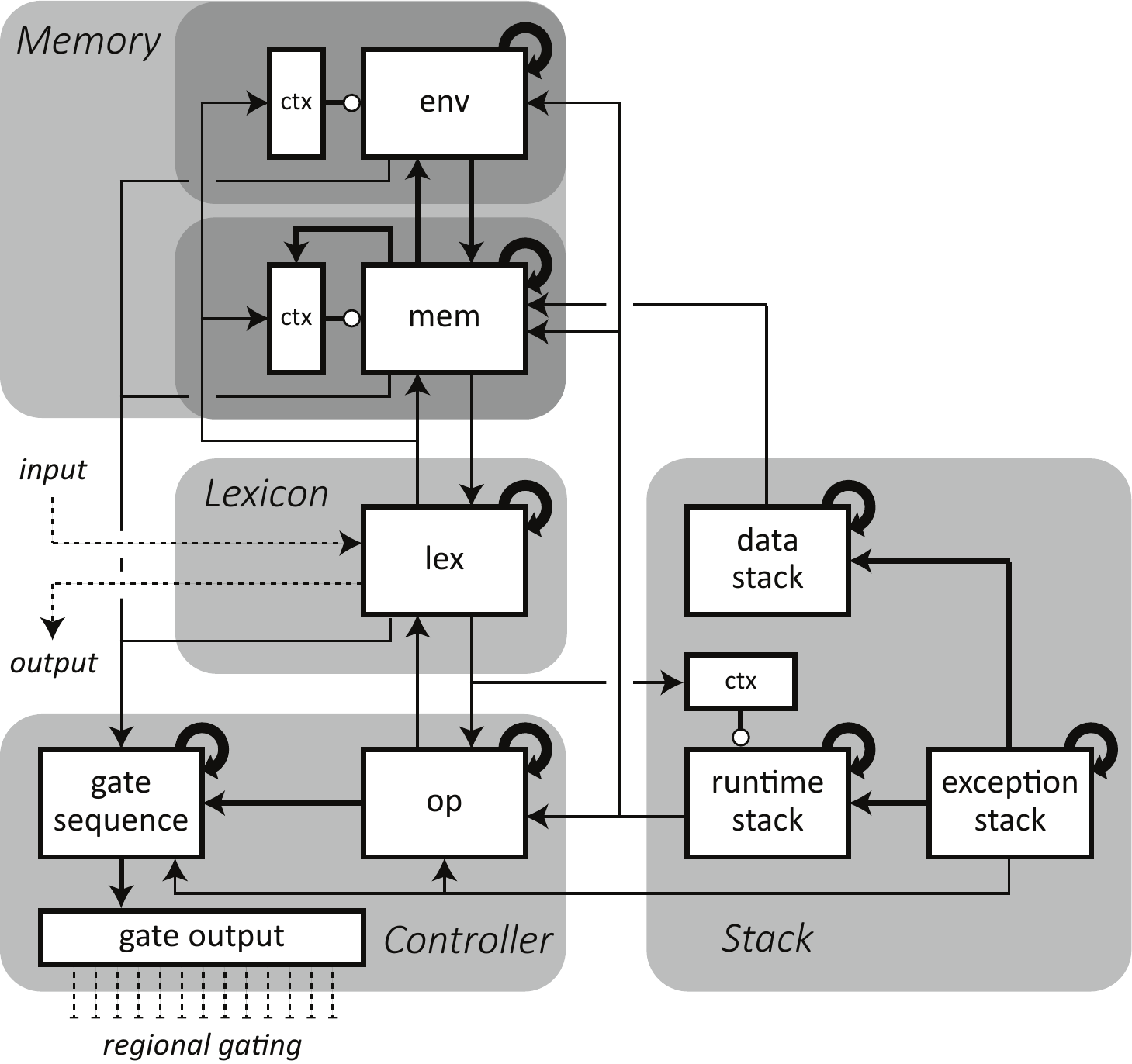}
\end{center}
\caption{ NeuroCERIL's neurocognitive architecture that learns to perform hypothetico-deductive causal inference. This architecture is an extension of NeuroLISP \citep{davis2022neurolisp}, and is made up of several recurrent neural regions (boxes) with inter-regional connections (solid arrows) that are divided into sub-networks (grey background boxes). Like NeuroLISP, NeuroCERIL implements an interpreter for a LISP-like programming language that is used to implement high-level algorithms. Programs and other data are stored as systems of learned attractors in the \texttt{mem} region (center), and are evaluated via top-down control of connection gates (\textit{regional gating}, bottom left). Inputs and outputs to the model are mediated by the \texttt{lex} region (center), which represents symbols as distributed patterns of activity that can be dynamically associated with activity patterns in adjacent regions (e.g., data structures in \texttt{mem}). NeuroCERIL implements a new class system using existing circuitry for variable bindings (connectivity between the \texttt{mem} and \texttt{env} regions, top), and also includes a new \texttt{exception stack} region (bottom right) that supports exception handling. These new features allow for efficient implementation of the causal inference algorithms described in this paper (see text for details). }\label{fig:neuroceril}
\end{figure*}

NeuroCERIL's architecture (shown in Figure \ref{fig:neuroceril}) is an extension of NeuroLISP, a programmable neural network that learns to store and evaluate programs written in a subset of the Common LISP programming language.
Many of the details of NeuroCERIL's functionality are shared with NeuroLISP and can be found in \citep{davis2022neurolisp}.
Here we provide a brief overview and highlight the novel features of NeuroCERIL's architecture that extend its computational capabilities beyond NeuroLISP.

Like NeuroLISP, NeuroCERIL represents programs and other symbolic data structures as learned systems of dynamical attractor states and associative transitions between attractor states \citep{davis2021compositional}.
Programs are evaluated via top-down control of gated connectivity between and within neural regions. This guides the flow of activity according to instructions retrieved from neural memory, much like a conventional computer architecture controls data flow according to instruction opcodes.
Importantly, NeuroCERIL also controls its own learning in this way, allowing it to construct, access, and modify data structures stored in memory during program evaluation.
Inputs and outputs are mediated by gated connectivity between the outer environment and a special region that represents discrete symbols as unique patterns of activity (\texttt{lex}, center of Figure \ref{fig:neuroceril}).
These connections allow NeuroCERIL to read symbolic inputs, including representations of programs, and output the results of program evaluation.
During imitation learning, a demonstration recorded in SMILE is provided as a sequence of symbolic inputs, and NeuroCERIL outputs a sequence of symbolic outputs that encodes the inferred causal explanation.

NeuroCERIL is initialized with a learned program-independent virtual machine composed of procedures that implement the primitive operations of its programming language \citep{katz2019programmable}.
After initialization, NeuroCERIL is programmed with the causal inference algorithm described in Section \ref{sec3.2}, which is expressed in the language of NeuroCERIL's virtual machine.
The details of initialization and program learning can be found in \citep{davis2022neurolisp}.

NeuroCERIL's virtual machine supports two major innovations that extend its computational capabilities beyond NeuroLISP and ease the implementation of its causal inference algorithm: a class system and an exception handling system.
The class system allows specification of reusable programs (i.e., class methods) for initializing and modifying instances of complex data structures such as causal hypotheses, cause-effect knowledge, and observed actions.
Instances of classes, called objects, are stored as collections of named pointers to other memories (i.e., class attributes).
The underlying implementation of objects makes use of the existing mechanisms for variable binding in NeuroCERIL's virtual machine; objects have corresponding lexical namespaces that store attributes as variable bindings (see \citep{davis2022neurolisp} for details on variable binding in NeuroLISP).

Exceptions are errors that occur during program evaluation, and are triggered by events such as attempted access to undefined variables, attributes, or class methods.
The exception handling system provides a mechanism for specifying dynamic responses to exceptions.
This obviates the need for excessive program expressions that perform checks on data before access; a program instead can specify what should be done if retrieval fails.
For example, when evoking hypotheses to explain an observed action, NeuroCERIL consults its causal knowledge-base to retrieve cause-effect recipes that are relevant to the observed action (see Section \ref{sec3.2}).
If the knowledge-base does not contain any entry for the observed action type, retrieval will result in an exception that can be easily handled by skipping the evocation process.

Exception handling is supported by the \texttt{exception stack} region (bottom right of Figure \ref{fig:neuroceril}), which maintains pointers to activity states in other regions that represent the state of the virtual machine.
This region functions like the \texttt{runtime} and \texttt{data stack} regions (shared with NeuroLISP), which represent stack frames as distributed patterns of activity that have learned associations with activity patterns in other regions \citep{davis2022neurolisp}.
Responses to exceptions are specified in programs with ``try" expressions that include a primary sub-expression to evaluate, and an additional sub-expression representing the response.
When a ``try" expression is evaluated, the virtual machine first stashes its state on the exception stack, which involves learning associations in the pathways exiting the \texttt{exception stack} region.
Then, the virtual machine attempts to evaluate the primary sub-expression.
If an exception occurs, the virtual machine retrieves its prior state from the exception stack, and evaluates the response sub-expression.
Upon completion, the top of the exception state is popped, and evaluation of the program continues.

\subsection{Experimental Evaluation}\label{sec3.4}

To evaluate NeuroCERIL, we performed empirical experiments using a battery of test demonstrations that was used to test CERIL.
These tests include procedural maintenance tasks involving replacing, swapping, and discarding mock hard drives in a docking assembly, as well as toy block stacking tasks (see \citep{katz2017novel} for details).
We compared NeuroCERIL's output with CERIL's to confirm that it performs comparably, and carried out additional analysis on its memory usage and runtime to determine how well it scales with the length of demonstrations.

Runtime was measured as the number of timesteps in model simulations, and memory usage was evaluated by monitoring each simulation to count the number of associations that were learned during causal inference.
Specifically, we monitored learning of attractor states and transitions in the underlying neural networks (stored in the recurrent connectivity of the \texttt{mem} region in Figure \ref{fig:neuroceril}), as well as associations between namespaces and memory states that represent variable bindings for both local variables and object attributes (stored in the connection from \texttt{env} to \texttt{mem} in Figure \ref{fig:neuroceril}).
These associations represent the core data structures used during causal inference, such as observed actions, hypotheses, and inferred causes.
We report the associations formed specifically during the inference process, and exclude those that represent the causal inference programs and cause-effect knowledge that is shared across demonstrations.

We further examined NeuroCERIL's memory access patterns to gain a better understanding of its memory usage.
We hypothesized that the majority of memories constructed during inference would be highly transient memories that are only accessed across brief intervals of time, such as abandoned causal hypotheses.
This would indicate that NeuroCERIL might benefit from a functionally distinct short-term memory system in which memories rapidly fade if they are not refreshed by retrieval, much like human working memory.
To test this hypothesis, we recorded instances of memory construction and access during inference, excluding demonstration-independent memories such as program representations and cause-effect knowledge.
For each recorded memory, we determined its ``lifespan" as the interval between its initial learning and the final time it was retrieved during the simulation (i.e., a memory is ``born" when it is first learned, and ``dies" after its last retrieval during the simulation).
We then calculated the number of ``living" memories over the course of the inference process and compared it to the total number of memories constructed.
This provides a metric for the proportion of memories that are being actively utilized for causal inference.


\section{Results}\label{sec4}



\begin{figure*}[!t]
\begin{center}
\includegraphics[width=0.87\textwidth]{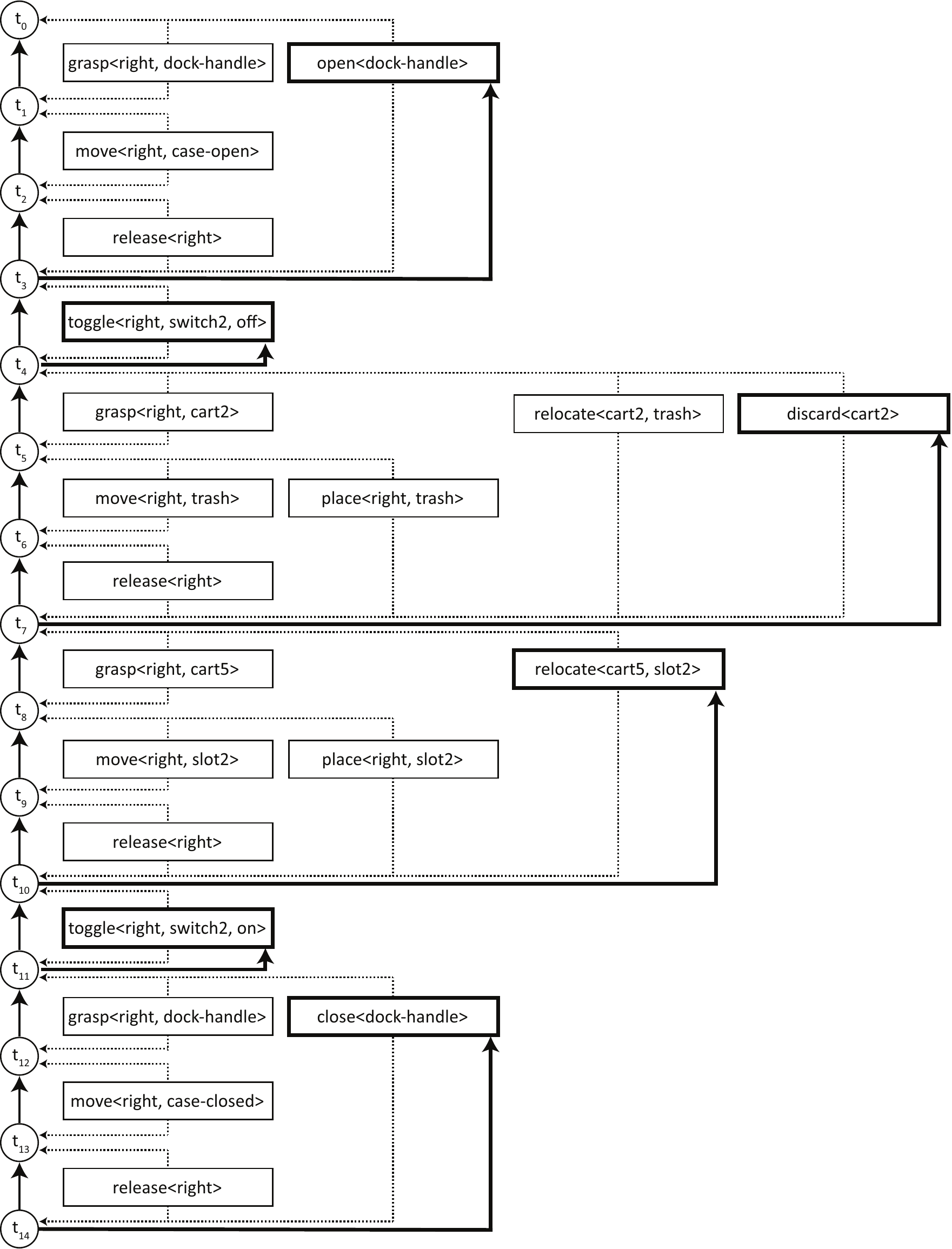}
\end{center}
\caption{ Causal interpretation produced by NeuroCERIL for the \textit{replace red with spare (1)} task, which involves replacing a broken disk cartridge (\textit{cart2}) in a mock disk drive drawer with a fresh cartridge (\textit{cart5}). Actions and causal intentions are represented by rectangles that indicate the type of action/intention along with its parameters. Each action/intention points to its start and end timepoints, represented by circles (left side), which delineate concrete observed actions (leftmost column of boxes). The top-level explanation, composed mostly of abstract intentions, is indicated by bold boxes. NeuroCERIL reconstructs this explanation by following the shortest path from the final ($t_{14}$) to initial ($t_0$) timepoints using parsimony pointers (bold arrows, shown only for relevant timepoints; see Section \ref{sec3.2}).
}\label{fig:sample_demo}
\end{figure*}

Table \ref{table:battery} shows the results for the same benchmark battery of procedural maintenance task demonstrations that were used to verify CERIL's functionality.
For each task, we report the number of actions recorded in the demonstration (\textbf{Act}), the number of top-level intentions in NeuroCERIL's causal interpretation (\textbf{Interp}), the number of timesteps of neural network simulation required for causal inference (\textbf{Timesteps}), and three measurements of learned associations that indicate model memory usage: the number of learned attractors (\textbf{Attr}) and attractor transitions (\textbf{Transit}) in the \texttt{mem} region, and the number of learned variable bindings (\textbf{Bindings}).
NeuroCERIL produced causal interpretations (sequences of top-level intentions) equivalent to the minimum cardinality explanations identified by CERIL for each of the tests. Figure \ref{fig:sample_demo} shows an example of the causal interpretation inferred for the \textit{replace red with spare (1)} task.

\begin{table*}[!h]
\caption{ NeuroCERIL performance on battery of robotic imitation learning tasks }
\begin{center}
\begin{tabular}{| l | r | r | r | r | r | r |}
\hline \textbf{Demonstrated Task} & \textbf{Act} & \textbf{Interp} & \textbf{Timesteps} & \textbf{Attr} & \textbf{Transit} & \textbf{Bindings} \\
\hline Remove red drive (1)       &  7 &  3 &  353825 & 197 & 362 & 661 \\
\hline Remove red drive (2)       & 10 &  4 &  490085 & 250 & 468 & 917 \\
\hline Replace red with spare (1) & 14 &  6 &  653758 & 341 & 642 & 1243 \\
\hline Replace red with spare (2) & 14 &  6 &  653758 & 341 & 642 & 1243 \\
\hline Replace red with green (1) & 15 &  7 &  668955 & 356 & 670 & 1276 \\
\hline Replace red with green (2) & 15 &  7 &  668955 & 356 & 670 & 1276 \\
\hline Swap red with green (1)    & 16 &  8 &  668889 & 357 & 672 & 1278 \\
\hline Swap red with green (2)    & 16 &  8 &  668981 & 361 & 680 & 1285 \\
\hline Toy blocks (IL)            & 24 &  8 & 1224377 & 591 & 1150 & 2326 \\
\hline Toy blocks (AI)            & 30 & 10 & 1524253 & 735 & 1434 & 2905 \\
\hline Toy blocks (UM)            & 39 & 13 & 1975927 & 945 & 1848 & 3763 \\
\hline 
\end{tabular}
\end{center}
\label{table:battery}
\end{table*}


Runtime and memory usage results provide an empirical estimate of the complexity of NeuroCERIL's hypothetico-deductive causal inference algorithm.
Figure \ref{fig:results_complexity} shows runtime and memory usage relative to the length of input demonstrations (\textbf{Act} in Table \ref{table:battery}).
Each datapoint corresponds to a row in Table \ref{table:battery}, and the dashed lines show the results of linear regression computed for each metric.
These results can be compared to Table 1 in \citep{katz2017novel}%
\footnote{We used slightly more complex versions of the IL and AI block stacking tasks that include more blocks and actions than those reported in \citep{katz2017novel}.}%
, as well as the theoretical analysis of CERIL's complexity.
Whereas CERIL exhibits a super-linear scaling of runtime and memory usage (indicated by the number of recognized top-level covers), NeuroCERIL's runtime and memory usage scale linearly with the length of the demonstration.
This is due to its online processing of demonstrations and incremental updating of data structures in memory that implicitly represent possible explanations.

\pgfplotstableread{data_complexity.txt}\complexitytable

\pgfplotstableread{data_lifespans.txt}\lifespantable

\pgfplotstablegetrowsof{\lifespantable}
\pgfmathsetmacro\numrows{\pgfplotsretval-1}

\pgfplotstableread{data_memory_load.txt}\memoryloadtable

\begin{figure}[!t]
\begin{minipage}[c]{0.475\linewidth}
\begin{center}
    \begin{subfigure}[b]{\linewidth}
    \centering
    \caption{Memory\ Usage}
    \begin{tikzpicture}[trim axis left, trim axis right]
    \selectcolormodel{gray}
    \begin{axis}[
        ytick distance=500,
        xtick pos=bottom,
        ytick pos=left,
        ymin=0,
        ymax=4096,
        xmin=5,
        xmax=40,
    	xlabel=Demonstration\ Length,
    	ylabel=Count,
    	width=\columnwidth,
    	height=1.3\columnwidth,
        legend style={at={(0.1,0.95)},anchor=north west},
        ]
        
\addlegendimage{empty legend}
    
    \addplot[
        scatter,
        black,
        nodes near coords,
        only marks,
        point meta=explicit symbolic,
        mark=triangle,
    ] table[x=N, y=B] {\complexitytable};
    
    \addplot[
        scatter,
        black,
        nodes near coords,
        only marks,
        point meta=explicit symbolic,
        mark=x,
    ] table[x=N, y=S] {\complexitytable};
    
    \addplot[
        scatter,
        black,
        nodes near coords,
        only marks,
        point meta=explicit symbolic,
        mark=o,
    ] table[x=N, y=A] {\complexitytable};

    \addplot [dashed, red] table[
        x=N, y={create col/linear regression={y=A}}
    ] {\complexitytable};
    \addplot [dashed, red] table[
        x=N, y={create col/linear regression={y=S}}
    ] {\complexitytable};
    \addplot [dashed, red] table[
        x=N, y={create col/linear regression={y=B}}
    ] {\complexitytable};

\addlegendentry{\hspace{-.4cm}\textbf{Memory Type}}
\addlegendentry{Bindings}
\addlegendentry{Transitions}
\addlegendentry{Attractors}
\end{axis}
\end{tikzpicture}%
\end{subfigure}%

\begin{subfigure}[b]{\linewidth}
    \centering
    \caption{Runtime}
    \begin{tikzpicture}[trim axis left, trim axis right]
    \selectcolormodel{gray}
    \begin{axis}[
        ytick distance=500000,
        ytick={0, 500000, 1000000, 1500000, 2000000},
        yticklabels={0, 500k, 1m, 1.5m, 2m},
        scaled y ticks=false,
        xtick pos=bottom,
        ytick pos=left,
        ymin=0,
        ymax=2250000,
        xmin=5,
        xmax=40,
    	xlabel=Demonstration\ Length,
    	ylabel=Timesteps,
    	width=\columnwidth,
    	height=0.9\columnwidth,
        legend style={at={(0.1,0.9)},anchor=north west},
        ]
        
    \addplot[
        scatter,
        black,
        nodes near coords,
        only marks,
        point meta=explicit symbolic,
        mark=star,
    ] table[x=N, y=T] {\complexitytable};

    \addplot [dashed, red] table[
        x=N, y={create col/linear regression={y=T}}
    ] {\complexitytable};
    
\end{axis}
\end{tikzpicture}%
\end{subfigure}%
\end{center}
\caption{ NeuroCERIL's memory usage and runtime during causal inference. Each data point corresponds to an individual imitation learning test task (rows in Table \ref{table:battery}). Dashed lines are lines of best fit computed with linear regression, and show that memory usage and runtime scale linearly with the length of the demonstration (x-axis). \textbf{(a)} Memory usage is reported as the number of learned attractor states, attractor transitions, and variable bindings generated during causal inference (see text for details). \textbf{(b)} Runtime is reported as the number of timesteps of neural model simulation required for causal inference. }
\label{fig:results_complexity}
\end{minipage}
\hfill
\begin{minipage}[c]{0.475\linewidth}
\begin{center}
    \begin{subfigure}[b]{\linewidth}
    \centering
    \caption{Memory\ Lifespans}
    \begin{tikzpicture}[trim axis left, trim axis right]
    \selectcolormodel{gray}
    \begin{axis}[
        xtick pos=bottom,
        ytick pos=left,
        ymin=0,
        ymax=349,
        xmin=-100,
        xmax=5500,
    	xlabel=Time\ (Activation\ Index),
    	ylabel=Memory\ Index,
    	width=\columnwidth,
    	height=\columnwidth,
        legend style={at={(0.1,0.9)},anchor=north west},
        ]

    \pgfplotsinvokeforeach {0,...,\numrows}{
       \pgfplotstablegetelem{#1}{First}\of\lifespantable
       \pgfmathsetmacro{\first}{\pgfplotsretval}
       \pgfplotstablegetelem{#1}{Last}\of\lifespantable
       \pgfmathsetmacro{\last}{\pgfplotsretval}
       
       \addplot+[solid,color=blue,mark=none] coordinates {(\first,#1) (\last,#1)};
    }

\end{axis}
\end{tikzpicture}%
\end{subfigure}%

\begin{subfigure}[b]{\linewidth}
    \centering
    \caption{Memory\ Load}
    \begin{tikzpicture}[trim axis left, trim axis right]
    \selectcolormodel{gray}
    \begin{axis}[
        xtick pos=bottom,
        ytick pos=left,
        ymin=0,
        ymax=349,
        xmin=-100,
        xmax=5500,
    	xlabel=Time\ (Activation\ Index),
    	ylabel=Count,
    	width=\columnwidth,
    	height=0.85\columnwidth,
        legend style={at={(0.1,0.9)},anchor=north west},
        ]

\addlegendimage{empty legend}
    \addplot[black] table[x=Index, y=Total] {\memoryloadtable};
    \addplot[orange] table[x=Index, y=Active] {\memoryloadtable};

\addlegendentry{\hspace{-.4cm}\textbf{}}
\addlegendentry{Total}
\addlegendentry{``Alive"}
\end{axis}
\end{tikzpicture}%
\end{subfigure}%
\end{center}
\caption{ ``Lifespans" of memory attractors constructed during causal inference on the \textit{replace red with spare (1)} task, reported as the interval between initial learning and final retrieval. The x-axis indexes model simulation timesteps in which an attractor is learned or retrieved. \textbf{(a)} Each horizontal line represents the lifespan of one memory attractor, indexed along the y-axis. Shorter lines indicate that a memory attractor is only accessed over a brief interval, while longer lines indicate memories that are utilized over longer periods of time. \textbf{(b)} Memory load, reported as the total number of memories learned over time compared to the number of ``living" memory attractors (i.e., attractors that have been learned at or before a given time and will be retrieved at a later time). While the total number of memories steadily increases over time, the majority of these memories are rapidly abandoned, and are only accessed over a brief period of time. }
\label{fig:results_lifespans}
\end{minipage}
\end{figure}

In further analysis of memory usage, we focus on learned memory attractors, as they are a bottleneck for neural attractor memory \citep{davis2021compositional}.
We present the results for the \textit{replace red with spare (1)} demonstration, but note that the results for other demonstrations are comparable.
Figure \ref{fig:results_lifespans}a shows the ``lifespans" of memory attractors constructed during causal inference for this demonstration, computed as the interval between initial learning and final retrieval.
The x-axis indexes timesteps in which a memory attractor is constructed or retrieved, and each horizontal line indicates the lifespan of one memory attractor, indexed along the y-axis.
Some memories remain alive through the majority of the inference process, such as representations of the environment and inferred causes that make up the final top-level cover.
Others have relatively short lifespans, such as falsified causal hypotheses.
We refer to the ``living memories" at a given timestep as the set of memories that have been learned prior to that timestep, and that will be accessed at a later timestep (i.e., a memory ``dies" after the final timestep in which it is accessed).
Figure \ref{fig:results_lifespans}b shows the total number of memory attractors learned over the course of the inference process, along with the number of ``living" memories at each timestep, which corresponds to the number of overlapping horizontal lines at each point along the x-axis in Figure \ref{fig:results_lifespans}a.
Although the total number of learned memories increases steadily over time, the majority of these memories have relatively short lifespans.
As a result, the number of ``living" memories remains fairly stable over time, and never exceeds 20\% of the total learned memories.




\section{Discussion}\label{sec5}

In this paper, we presented NeuroCERIL, a brain-inspired neurocognitive controller for social robots that learn procedural tasks from human-provided demonstrations (i.e., robotic imitation learning).
NeuroCERIL infers the intentions underlying demonstrated behavior using a novel causal inference algorithm based on human-like hypothetico-deductive reasoning, which combines bottom-up abductive inference with top-down predictive verification.
This approach allows NeuroCERIL to iteratively construct plausible interpretations of demonstrated behavior as it is observed, make verifiable predictions about subsequent behavior, and generate compact explanations in terms of abstract intentions that can be generalized to novel environments.
We evaluated NeuroCERIL on a benchmark battery of procedural maintenance and toy block-stacking tasks recorded in a virtual environment, demonstrating that it works effectively in robotic imitation learning domains.
Our empirical results also show that the model scales well with the length of demonstrated action sequences, and that the majority of its memory usage during causal inference is dedicated to transient short-term memories, much like human working memory.

NeuroCERIL is distinguished from prior approaches to robotic imitation learning by its use of neural computations to understand demonstrated behavior in terms of causal relations that are directly related to high-level planning and cognitive-motor control.
This not only affords generalization during imitation, but also facilitates an understanding of roles and perspectives that is critical to human-robot collaboration \citep{trafton2005enabling}.
In addition, NeuroCERIL maintains a model of the external environment in memory and tracks changes that are induced by demonstrated motor activity.
NeuroCERIL's understanding of demonstrations therefore provides an awareness of the physical consequences of behavior that is critical for safe and effective deployment of robots in sensitive environments.

Causal reasoning and compositionality are widely considered to be critical components of human cognition that are challenging for contemporary neural models to learn \citep{lake2017building, hupkes2020compositionality, lake2018generalization, loula2018rearranging}.
NeuroCERIL performs causal reasoning with compositional models in working memory that represent the external environment and encode high-level behavioral plans, and is therefore a significant step toward developing neural networks with human-like reasoning capabilities.
In addition, we have previously proposed that neural models of working memory control, particularly in humanoid robots, provide a promising avenue to understanding conscious cognitive processing and its underlying basis in neural computations \citep{reggia2018humanoid, reggia2019modeling}.
NeuroCERIL is therefore also relevant to investigations of consciousness in machines and biological agents because it implements human-like cognitive algorithms in a brain-inspired neural architecture.

NeuroCERIL has several important limitations that suggest directions for future research.
In this paper, we have focused on the causal inference component of imitation learning (left side of Figure \ref{fig:ceril}), and have not addressed the generation of motor plans to implement learned skills during imitation.
In prior work, we have shown that programmable neural networks can implement basic hierarchical planning, and can perform adaptable motor control in simulated robots \citep{davis2021compositional, katz2021tunable}.
It is therefore feasible to integrate NeuroCERIL with low-level neural models of perception and motor control to create a complete neurocognitive imitation learning system that performs both causal inference and plan generation.

Our proposed hypothetico-deductive causal reasoning algorithm relies on constraints in demonstrated behavior.
In particular, implementations of abstract intentions must be performed in a fixed order as specified in the causal knowledge-base, and cannot be broken up by unrelated actions.
In reality, procedural tasks might involve interleaved action sequences performed with both hands, and may include steps that can be performed in arbitrary arrangements.
Thus, future work might involve modifying our causal inference algorithm to support these variations.
This might also permit generalization to additional cognitive domains in which hypothetico-deductive reasoning is relevant, such as visual scene understanding and linguistic processing.

Finally, NeuroCERIL uses a unified memory system that does not include functionally distinct short-term and long-term memory.
This means that long-term memories such as programs and causal knowledge may be gradually degraded as new short-term memories are constructed during program evaluation.
Our empirical results show that the majority of memories constructed during causal inference are only accessed during a narrow window of time, and are therefore highly transient short-term memories.
This suggests that NeuroCERIL would benefit from a functional separation of short-term and long-term memory to protect the latter from interference.

\section*{Acknowledgements}
This work was supported by ONR award N00014-19-1-2044.

\section*{Data Availability}
The datasets generated during and/or analysed during the current study are available in the NeuroCERIL repository, https://github.com/vicariousgreg/neuroceril

\bibliographystyle{unsrtnat}
\bibliography{bibliography}  

\newcommand{\noop}[1]{}
\begin{thebibliography}{41}
\providecommand{\natexlab}[1]{#1}
\providecommand{\url}[1]{\texttt{#1}}
\expandafter\ifx\csname urlstyle\endcsname\relax
  \providecommand{\doi}[1]{doi: #1}\else
  \providecommand{\doi}{doi: \begingroup \urlstyle{rm}\Url}\fi

\bibitem[Jones(2009)]{jones2009development}
Susan~S Jones.
\newblock The development of imitation in infancy.
\newblock \emph{Philosophical Transactions of the Royal Society B: Biological
  Sciences}, 364\penalty0 (1528):\penalty0 2325--2335, 2009.

\bibitem[Meltzoff et~al.(2009)Meltzoff, Kuhl, Movellan, and
  Sejnowski]{meltzoff2009foundations}
Andrew~N Meltzoff, Patricia~K Kuhl, Javier Movellan, and Terrence~J Sejnowski.
\newblock Foundations for a new science of learning.
\newblock \emph{Science}, 325\penalty0 (5938):\penalty0 284--288, 2009.

\bibitem[Ravichandar et~al.(2020)Ravichandar, Polydoros, Chernova, and
  Billard]{ravichandar2020recent}
Harish Ravichandar, Athanasios~S Polydoros, Sonia Chernova, and Aude Billard.
\newblock Recent advances in robot learning from demonstration.
\newblock \emph{Annual Review of Control, Robotics, and Autonomous Systems},
  3:\penalty0 297--330, 2020.

\bibitem[Hussein et~al.(2017)Hussein, Gaber, Elyan, and
  Jayne]{hussein2017imitation}
Ahmed Hussein, Mohamed~Medhat Gaber, Eyad Elyan, and Chrisina Jayne.
\newblock Imitation learning: A survey of learning methods.
\newblock \emph{ACM Computing Surveys (CSUR)}, 50\penalty0 (2):\penalty0 1--35,
  2017.

\bibitem[Billard et~al.(2008)Billard, Calinon, Dillmann, and
  Schaal]{billard2008survey}
Aude Billard, Sylvain Calinon, Ruediger Dillmann, and Stefan Schaal.
\newblock Survey: Robot programming by demonstration.
\newblock Technical report, Springer, 2008.

\bibitem[Schaal(1999)]{schaal1999imitation}
Stefan Schaal.
\newblock Is imitation learning the route to humanoid robots?
\newblock \emph{Trends in cognitive sciences}, 3\penalty0 (6):\penalty0
  233--242, 1999.

\bibitem[Trafton et~al.(2005)Trafton, Cassimatis, Bugajska, Brock, Mintz, and
  Schultz]{trafton2005enabling}
J~Gregory Trafton, Nicholas~L Cassimatis, Magdalena~D Bugajska, Derek~P Brock,
  Farilee~E Mintz, and Alan~C Schultz.
\newblock Enabling effective human-robot interaction using perspective-taking
  in robots.
\newblock \emph{IEEE Transactions on Systems, Man, and Cybernetics-Part A:
  Systems and Humans}, 35\penalty0 (4):\penalty0 460--470, 2005.

\bibitem[Katz et~al.(2017)Katz, Huang, Hauge, Gentili, and
  Reggia]{katz2017novel}
Garrett Katz, Di-Wei Huang, Theresa Hauge, Rodolphe Gentili, and James Reggia.
\newblock A novel parsimonious cause-effect reasoning algorithm for robot
  imitation and plan recognition.
\newblock \emph{IEEE Transactions on Cognitive and Developmental Systems},
  10\penalty0 (2):\penalty0 177--193, 2017.

\bibitem[Davis et~al.(2022)Davis, Katz, Gentili, and
  Reggia]{davis2022neurolisp}
Gregory~P Davis, Garrett~E Katz, Rodolphe~J Gentili, and James~A Reggia.
\newblock Neurolisp: High-level symbolic programming with attractor neural
  networks.
\newblock \emph{Neural Networks}, 146:\penalty0 200--219, 2022.

\bibitem[Bandura(2017)]{bandura2017psychological}
Albert Bandura.
\newblock \emph{Psychological modeling: Conflicting theories}.
\newblock Transaction Publishers, New Jersey, USA, 2017.

\bibitem[Meltzoff(1995)]{meltzoff1995understanding}
Andrew~N Meltzoff.
\newblock Understanding the intentions of others: re-enactment of intended acts
  by 18-month-old children.
\newblock \emph{Developmental psychology}, 31\penalty0 (5):\penalty0 838, 1995.

\bibitem[Baldwin and Baird(2001)]{baldwin2001discerning}
Dare~A Baldwin and Jodie~A Baird.
\newblock Discerning intentions in dynamic human action.
\newblock \emph{Trends in cognitive sciences}, 5\penalty0 (4):\penalty0
  171--178, 2001.

\bibitem[Tomasello et~al.(1993)Tomasello, Kruger, and
  Ratner]{tomasello1993cultural}
Michael Tomasello, Ann~Cale Kruger, and Hilary~Horn Ratner.
\newblock Cultural learning.
\newblock \emph{Behavioral and brain sciences}, 16\penalty0 (3):\penalty0
  495--511, 1993.

\bibitem[Oztop et~al.(2013)Oztop, Kawato, and Arbib]{oztop2013mirror}
Erhan Oztop, Mitsuo Kawato, and Michael~A Arbib.
\newblock Mirror neurons: functions, mechanisms and models.
\newblock \emph{Neuroscience letters}, 540:\penalty0 43--55, 2013.

\bibitem[Jackson et~al.(2006)Jackson, Meltzoff, and Decety]{jackson2006neural}
Philip~L Jackson, Andrew~N Meltzoff, and Jean Decety.
\newblock Neural circuits involved in imitation and perspective-taking.
\newblock \emph{Neuroimage}, 31\penalty0 (1):\penalty0 429--439, 2006.

\bibitem[Fogassi et~al.(2005)Fogassi, Ferrari, Gesierich, Rozzi, Chersi, and
  Rizzolatti]{fogassi2005parietal}
Leonardo Fogassi, Pier~Francesco Ferrari, Benno Gesierich, Stefano Rozzi,
  Fabian Chersi, and Giacomo Rizzolatti.
\newblock Parietal lobe: from action organization to intention understanding.
\newblock \emph{Science}, 308\penalty0 (5722):\penalty0 662--667, 2005.

\bibitem[K{\"o}ster et~al.(2020)K{\"o}ster, Langeloh, Kliesch, Kanngiesser, and
  Hoehl]{koster2020motor}
Moritz K{\"o}ster, Miriam Langeloh, Christian Kliesch, Patricia Kanngiesser,
  and Stefanie Hoehl.
\newblock Motor cortex activity during action observation predicts subsequent
  action imitation in human infants.
\newblock \emph{NeuroImage}, 218:\penalty0 116958, 2020.

\bibitem[Reggia et~al.(2018)Reggia, Katz, and Davis]{reggia2018humanoid}
James~A Reggia, Garrett~E Katz, and Gregory~P Davis.
\newblock Humanoid cognitive robots that learn by imitating: Implications for
  consciousness studies.
\newblock \emph{Frontiers in Robotics and AI}, 5:\penalty0 1, 2018.

\bibitem[Huang et~al.(2015)Huang, Katz, Langsfeld, Gentili, and
  Reggia]{huang2015virtual}
Di-Wei Huang, Garrett Katz, Joshua Langsfeld, Rodolphe Gentili, and James
  Reggia.
\newblock A virtual demonstrator environment for robot imitation learning.
\newblock In \emph{2015 IEEE International Conference on Technologies for
  Practical Robot Applications (TePRA)}, pages 1--6. IEEE, 2015.

\bibitem[Duan et~al.(2017)Duan, Andrychowicz, Stadie, Ho, Schneider, Sutskever,
  Abbeel, and Zaremba]{duan2017one}
Yan Duan, Marcin Andrychowicz, Bradly Stadie, Jonathan Ho, Jonas Schneider,
  Ilya Sutskever, Pieter Abbeel, and Wojciech Zaremba.
\newblock One-shot imitation learning.
\newblock In \emph{Proceedings of the 31st International Conference on Neural
  Information Processing Systems}, pages 1087--1098, 2017.

\bibitem[Liu et~al.(2018)Liu, Gupta, Abbeel, and Levine]{liu2018imitation}
YuXuan Liu, Abhishek Gupta, Pieter Abbeel, and Sergey Levine.
\newblock Imitation from observation: Learning to imitate behaviors from raw
  video via context translation.
\newblock In \emph{2018 IEEE International Conference on Robotics and
  Automation (ICRA)}, pages 1118--1125. IEEE, 2018.

\bibitem[Xu et~al.(2018)Xu, Nair, Zhu, Gao, Garg, Fei-Fei, and
  Savarese]{xu2018neural}
Danfei Xu, Suraj Nair, Yuke Zhu, Julian Gao, Animesh Garg, Li~Fei-Fei, and
  Silvio Savarese.
\newblock Neural task programming: Learning to generalize across hierarchical
  tasks.
\newblock In \emph{2018 IEEE International Conference on Robotics and
  Automation (ICRA)}, pages 3795--3802. IEEE, 2018.

\bibitem[Sun et~al.(2018)Sun, Noh, Somasundaram, and Lim]{sun2018neural}
Shao-Hua Sun, Hyeonwoo Noh, Sriram Somasundaram, and Joseph Lim.
\newblock Neural program synthesis from diverse demonstration videos.
\newblock In \emph{International Conference on Machine Learning}, pages
  4790--4799. PMLR, 2018.

\bibitem[Bunel et~al.(2018)Bunel, Hausknecht, Devlin, Singh, and
  Kohli]{bunel2018leveraging}
Rudy Bunel, Matthew Hausknecht, Jacob Devlin, Rishabh Singh, and Pushmeet
  Kohli.
\newblock Leveraging grammar and reinforcement learning for neural program
  synthesis.
\newblock \emph{arXiv preprint arXiv:1805.04276}, 2018.

\bibitem[Silver et~al.(2016)Silver, Huang, Maddison, Guez, Sifre, Van
  Den~Driessche, Schrittwieser, Antonoglou, Panneershelvam, Lanctot,
  et~al.]{silver2016mastering}
David Silver, Aja Huang, Chris~J Maddison, Arthur Guez, Laurent Sifre, George
  Van Den~Driessche, Julian Schrittwieser, Ioannis Antonoglou, Veda
  Panneershelvam, Marc Lanctot, et~al.
\newblock Mastering the game of go with deep neural networks and tree search.
\newblock \emph{Nature}, 529\penalty0 (7587):\penalty0 484--489, 2016.

\bibitem[Kalyan et~al.(2018)Kalyan, Mohta, Polozov, Batra, Jain, and
  Gulwani]{kalyan2018neural}
Ashwin Kalyan, Abhishek Mohta, Oleksandr Polozov, Dhruv Batra, Prateek Jain,
  and Sumit Gulwani.
\newblock Neural-guided deductive search for real-time program synthesis from
  examples.
\newblock \emph{arXiv preprint arXiv:1804.01186}, 2018.

\bibitem[Davis et~al.(2021)Davis, Katz, Gentili, and
  Reggia]{davis2021compositional}
Gregory~P. Davis, Garrett~E. Katz, Rodolphe~J. Gentili, and James~A. Reggia.
\newblock Compositional memory in attractor neural networks with one-step
  learning.
\newblock \emph{Neural Networks}, 138:\penalty0 78--97, 2021.
\newblock ISSN 0893-6080.

\bibitem[Katz et~al.(2019)Katz, Davis, Gentili, and
  Reggia]{katz2019programmable}
Garrett~E Katz, Gregory~P Davis, Rodolphe~J Gentili, and James~A Reggia.
\newblock A programmable neural virtual machine based on a fast store-erase
  learning rule.
\newblock \emph{Neural Networks}, 119:\penalty0 10--30, 2019.

\bibitem[Sylvester and Reggia(2016)]{sylvester2016engineering}
Jared Sylvester and James Reggia.
\newblock Engineering neural systems for high-level problem solving.
\newblock \emph{Neural Networks}, 79:\penalty0 37--52, 2016.

\bibitem[Katz et~al.(2021)Katz, Akshay, Davis, Gentili, and
  Reggia]{katz2021tunable}
Garrett~E Katz, .~Akshay, Gregory~P Davis, Rodolphe~J Gentili, and James~A
  Reggia.
\newblock Tunable neural encoding of a symbolic robotic manipulation algorithm.
\newblock \emph{Frontiers in Neurorobotics}, page 167, 2021.

\bibitem[Gentili et~al.(2015)Gentili, Oh, Huang, Katz, Miller, and
  Reggia]{gentili2015neural}
Rodolphe~J Gentili, Hyuk Oh, Di-Wei Huang, Garrett~E Katz, Ross~H Miller, and
  James~A Reggia.
\newblock A neural architecture for performing actual and mentally simulated
  movements during self-intended and observed bimanual arm reaching movements.
\newblock \emph{International Journal of Social Robotics}, 7\penalty0
  (3):\penalty0 371--392, 2015.

\bibitem[Lawson(2000{\natexlab{a}})]{lawson2000humans}
Anton~E Lawson.
\newblock How do humans acquire knowledge? and what does that imply about the
  nature of knowledge?
\newblock \emph{Science \& Education}, 9\penalty0 (6):\penalty0 577--598,
  2000{\natexlab{a}}.

\bibitem[Sprenger(2011)]{sprenger2011hypothetico}
Jan Sprenger.
\newblock Hypothetico-deductive confirmation.
\newblock \emph{Philosophy Compass}, 6\penalty0 (7):\penalty0 497--508, 2011.

\bibitem[Marcum(2012)]{marcum2012integrated}
James~A Marcum.
\newblock An integrated model of clinical reasoning: dual-process theory of
  cognition and metacognition.
\newblock \emph{Journal of evaluation in clinical practice}, 18\penalty0
  (5):\penalty0 954--961, 2012.

\bibitem[Reggia and Peng(1987)]{reggia1987modeling}
James~A Reggia and Yun Peng.
\newblock Modeling diagnostic reasoning: a summary of parsimonious covering
  theory.
\newblock \emph{Computer methods and programs in biomedicine}, 25\penalty0
  (2):\penalty0 125--134, 1987.

\bibitem[Lawson(2000{\natexlab{b}})]{lawson2000generality}
Anton~E Lawson.
\newblock The generality of hypothetico-deductive reasoning: Making scientific
  thinking explicit.
\newblock \emph{The American Biology Teacher}, 62\penalty0 (7):\penalty0
  482--495, 2000{\natexlab{b}}.

\bibitem[Lake et~al.(2017)Lake, Ullman, Tenenbaum, and
  Gershman]{lake2017building}
Brenden~M Lake, Tomer~D Ullman, Joshua~B Tenenbaum, and Samuel~J Gershman.
\newblock Building machines that learn and think like people.
\newblock \emph{Behavioral and brain sciences}, 40, 2017.

\bibitem[Hupkes et~al.(2020)Hupkes, Dankers, Mul, and
  Bruni]{hupkes2020compositionality}
Dieuwke Hupkes, Verna Dankers, Mathijs Mul, and Elia Bruni.
\newblock Compositionality decomposed: How do neural networks generalise?
\newblock \emph{Journal of Artificial Intelligence Research}, 67:\penalty0
  757--795, 2020.

\bibitem[Lake and Baroni(2018)]{lake2018generalization}
Brenden Lake and Marco Baroni.
\newblock Generalization without systematicity: On the compositional skills of
  sequence-to-sequence recurrent networks.
\newblock In \emph{International Conference on Machine Learning}, pages
  2873--2882, 2018.

\bibitem[Loula et~al.(2018)Loula, Baroni, and Lake]{loula2018rearranging}
Jo{\~a}o Loula, Marco Baroni, and Brenden Lake.
\newblock Rearranging the familiar: Testing compositional generalization in
  recurrent networks.
\newblock In \emph{Proceedings of the 2018 EMNLP Workshop BlackboxNLP:
  Analyzing and Interpreting Neural Networks for NLP}, pages 108--114, 2018.

\bibitem[Reggia et~al.(2019)Reggia, Katz, and Davis]{reggia2019modeling}
James~A Reggia, Garrett~E Katz, and Gregory~P Davis.
\newblock Modeling working memory to identify computational correlates of
  consciousness.
\newblock \emph{Open Philosophy}, 2\penalty0 (1):\penalty0 252--269, 2019.

\end{thebibliography}






\end{document}